\documentclass[letterpaper, 10 pt, conference]{ieeeconf}
\IEEEoverridecommandlockouts
\let\labelindent\relax
\usepackage{graphicx}
\usepackage{enumitem}
\usepackage{multirow}
\usepackage{algorithm}
\usepackage{algcompatible}
\usepackage{amsmath}
\usepackage{amssymb}
\usepackage{multirow}
\usepackage{hhline}
\usepackage{xcolor}

\newcommand{\R}{\mathbb{R}}
\newcommand{\barr}[1]{\bar{\mathbf{#1}}}

\usepackage{caption, subcaption}
\definecolor{lbrown}{HTML}{80471C}
\makeatletter
\renewcommand{\COMMENT}[1]{\Statex \hskip\ALG@thistlm   {\color{lbrown}\(\triangleright\)\textit{#1}}}
\makeatother

\renewcommand{\arraystretch}{1.2}

\usepackage[
	activate   = {true},
	protrusion = false,
	expansion  = true,
	kerning    = true,
	spacing    = true,
	tracking   = false,
	auto       = true,
	selected   = true,
	factor     = 2000,
	stretch    = 20,
	shrink     = 20,
]{microtype}

\usepackage[english=american]{csquotes}
\MakeOuterQuote{"}
\usepackage[
maxbibnames=99,
maxcitenames=2,
natbib=true,
style=numeric-comp,
backend=biber,
sorting=none,
giveninits=true,
url=false, 
doi=false,
eprint=false,
isbn=false,
]{biblatex}

\usepackage{filecontents}

\usepackage{xpatch}

\makeatletter         
\patchcmd\blx@bblinput{\blx@blxinit}
                      {\blx@blxinit
                      }{}{\fail}
\makeatother

\makeatletter
\let\NAT@parse\undefined
\makeatother
\usepackage[pdfa,colorlinks,bookmarksopen,bookmarksnumbered,allcolors=brown]{hyperref}
\usepackage[english]{babel}

\usepackage[nameinlink,capitalise]{cleveref}
\crefname{line}{line}{lines}
\crefname{figure}{Fig.}{Figs.}
\Crefname{figure}{Fig.}{Figs.}
\crefname{equation}{Eq.}{Eqs.}
\Crefname{equation}{Eq.}{Eqs.}
\crefname{section}{Sec.}{Secs.}
\Crefname{section}{Sec.}{Secs.}
\crefname{definition}{Def.}{Defs.}
\Crefname{definition}{Def.}{Defs.}
\crefname{algorithm}{Alg.}{Algs.}
\Crefname{algorithm}{Alg.}{Algs.}
\crefname{assumption}{Asm.}{Asms.}
\Crefname{assumption}{Asm.}{Asms.}
\crefname{theorem}{Thm.}{Thms.}
\Crefname{theorem}{Thm.}{Thms.}
\crefname{table}{Tbl.}{Tbls.}
\Crefname{table}{Tbl.}{Tbls.}
\crefname{subassumption}{Asm.}{Asms.}
\Crefname{subassumption}{Asm.}{Asms.}

\usepackage{tabularx}
\usepackage{multirow}
\usepackage{booktabs}
\usepackage{colortbl}
\usepackage{overpic}

\definecolor{lgray}{HTML}{F0F0F0}

\newcolumntype{I}{>{\centering}p{0.13\textwidth}}
\newcolumntype{Y}{>{\centering\arraybackslash}X}

\newcommand{\tblimgtext}[3]{\multirow{3}*{\vspace{-3pt}\hspace{-0.7em}\begin{overpic}[width=.15\textwidth]{table/#1}
      \put (2,2) {\rotatebox{90}{{\raggedleft\scalebox{0.8}{{#2}}}}}
      \put (84,2) {{{\raggedleft\textbf{#3}}}}
    \end{overpic}}}

\newcommand{\multicoltext}[3]{\multirow{3}*{\vspace{-3pt}\hspace{-0.7em}\begin{overpic}[width=.15\textwidth]{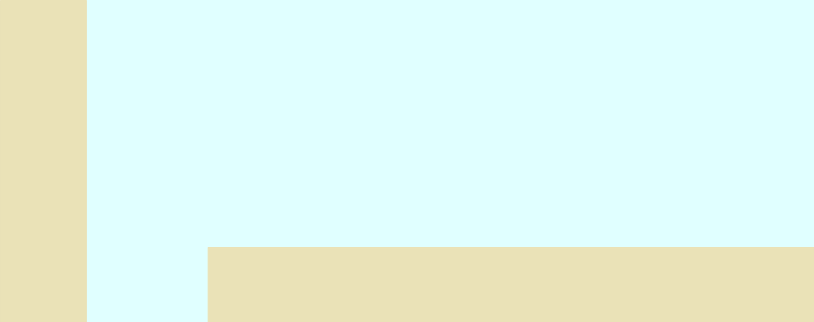}
      \put (2,2) {\rotatebox{90}{{\raggedleft\scalebox{0.8}{{#1}}}}}
      \put (34,17) {{{\raggedleft\textbf{#2}}}}
      \put (26,2) {{{{#3}}}}
    \end{overpic}}}

\usepackage{amsthm}

\newtheoremstyle{plaindefnobox}
  {}{}             %
  {}               %
  {}               %
  {\bfseries}      %
  {:}              %
  { }              %
  {\thmname{#1} \thmnumber{#2}: \thmnote{\em#3}} %
\theoremstyle{plaindefnobox}
\newtheorem{definition}{Manifold Definition}[]

\begin{document}
\title{\LARGE \bf Constrained Nonlinear Kaczmarz Projection on Intersections of Manifolds for Coordinated Multi-Robot Mobile Manipulation}

\author{Akshaya Agrawal$^{1}$, Parker Mayer$^{1}$, Zachary Kingston$^{2}$, and Geoffrey A. Hollinger$^{1}$
\thanks{$^{1}$Collaborative Robotics and Intelligent Systems (CoRIS) Institute, Oregon State University, \texttt{\{agrawaak, mayerpa, geoff.hollinger\} @oregonstate.edu}}
\thanks{$^{2}$Department of Computer Science, Purdue University \texttt{zkingston @purdue.edu}}%
\thanks{This work was supported by ONR award N0014-21-1-2052 and N00014-22-1-2114. Code is available at \url{https://github.com/JBVAkshaya/PlanningOnManifoldIntersection}}
}

\maketitle
\begin{abstract}
Cooperative manipulation tasks impose various structure-, task-, and robot-specific constraints on mobile manipulators. However, current methods struggle to model and solve these myriad constraints simultaneously. We propose a twofold solution: first, we model constraints as a family of manifolds amenable to simultaneous solving. Second, we introduce the constrained nonlinear Kaczmarz (cNKZ) projection technique to produce constraint-satisfying solutions. Experiments show that cNKZ dramatically outperforms baseline approaches, which cannot find solutions at all. We integrate cNKZ with a sampling-based motion planning algorithm to generate complex, coordinated motions for 3--6 mobile manipulators (18--36 DoF), with cNKZ solving up to 80 nonlinear constraints simultaneously and achieving up to a 92\% success rate in cluttered environments. We also demonstrate our approach on hardware using three Turtlebot3 Waffle Pi robots with OpenMANIPULATOR-X arms.
\end{abstract}

\section{Introduction}\label{section:introduction}
Autonomous mobile manipulator (AMM) teams can be employed for cooperative construction, mitigating risk to humans. AMM teams encounter mobility and task-specific constraints in coordinated transportation tasks. Consider a team of robots transporting a large solar panel to a space telescope amidst debris: the robots must move in tandem and are coupled by the carried structure. Thus, successful transportation requires satisfying diverse constraints (e.g., structure orientation, end-effector pose) simultaneously. This set of constraints also demands a unified representation encodable in a nonlinear optimizer to generate compliant configurations, which motion planners can use to create constrained trajectories.

\begin{figure}[t]
  \centering
  \includegraphics[width=8cm,trim={0cm 0cm 0cm 0cm},clip]{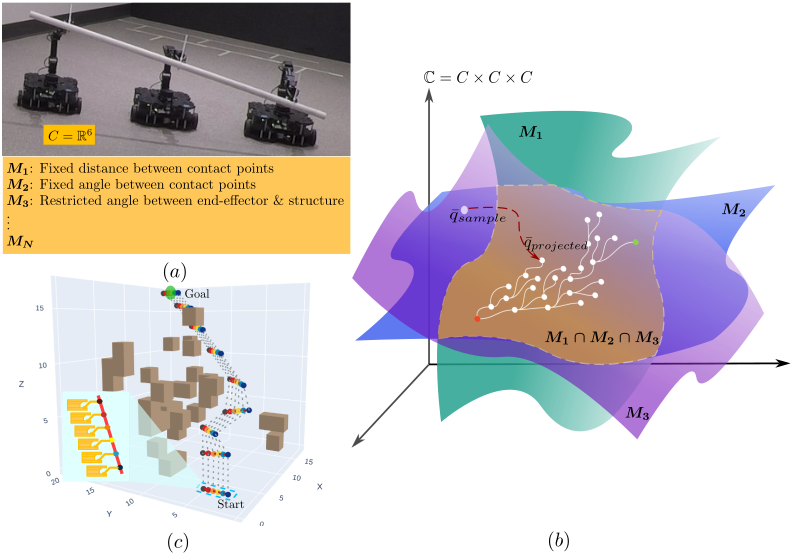}
  \caption{(a) Coordinated multi-robot problems require simultaneous constraint satisfaction. (b) We model constraints as a family of manifolds and propose a novel constrained nonlinear Kaczmarz (cNKZ) optimizer to find constraint-satisfying configurations. We integrate our approach within a sampling-based planner to find constraint-satisfying motion. (c) cNKZ solves problems for six robots (36 DoF) with 80 constraints in complex environments.}
  \label{figure:problem_formulation}
  \vspace{-1em}
\end{figure}

Existing approaches often prioritize a single constraint, addressing others as subordinate (e.g., in the null space of the first constraint) or treating the set of constraints as one composite function ~\cite{gammell2021asymptotically, gharbi2008sampling, cortes2005sampling, geraerts2004comparative, kingston2018sampling}. These approaches prove inadequate as constraint complexity grows, particularly with larger teams tackling intricate tasks  ~\cite{englert2021sampling, berensom2009manipulation}). The increasing numbers of constraints often result in substantial residuals, limiting the use of these methods in multi-robot coordinated planning~\cite{hauser2010multi}.

This paper presents the constrained nonlinear Kaczmarz (cNKZ) method for solving large, diverse systems of nonlinear constraints for coordinated multi-robot scenarios, e.g., cooperative mobile manipulation tasks. We propose encoding object geometry, task requirements, and robot limitations as manifolds to unify the constraint representation to support the use of cNKZ. We also incorporate cNKZ into a sampling-based motion planner, demonstrating on complex problems with over 30 DoF and more than 80 constraints. We assess computational efficiency in simulations with varied structures and increasing complexity and team sizes where cNKZ outperforms baseline approaches. Additionally, we demonstrate that our proposed planner can transport a linear structure in a cluttered environment on real hardware with three AMMs.

The main contributions of this work are:
\begin{itemize}
  \item A unified formulation of constraints representing structural, task-specific, and physical robot limitations as manifolds,
  \item Adaption of nonlinear Kaczmarz (NKZ) to constrained planning as a projection technique and introduce constrained NKZ (cNKZ), a modification with constraint-specific residual thresholds for solving large systems of constraints efficiently, and
  \item A centralized sampling-based constrained motion planning algorithm that integrates cNKZ to generate feasible paths for cooperative multi-robot manipulation.
\end{itemize}

\section{Related Work} \label{section:relatedwork}

\noindent\textbf{Constraint Definitions}\hspace{1em}%
\citet{zhang2021task} used task space regions (TSR) to define pose constraints for coordinated multi-robot transportation. However, current methods cannot solve multiple TSR chains simultaneously, only solving one while satisfying others through rejection strategies~\cite{berensom2009manipulation, berenson2011task}. Thus, non-serial, closed-chain, and task-specific constraints cannot be simultaneously solved using analytical or numerical methods. Other methods use implicit functions to define constraints for multi-robot task sequences (e.g.,~\cite{englert2021sampling}). This approach can model diverse constraints but is underdeveloped for coordinated manipulation. Inspired by prior work, we use implicit manifolds to define rigid structure, task-specific, and robot-specific constraints (\cref{section:manifold_model}) and leverage sampling-based planners to plan over intersections of manifolds (\cref{section:planning}). %

\noindent\textbf{Projection Methods}\hspace{1em}%
In motion planning domain, iterative nonlinear solvers like randomized gradient descent (RGD) and Newton's methods are effective projection techniques~\cite{akram2015newton, hao2021gradient}. However, Newton's method, which requires Jacobian inverse calculation, is limited to full-rank systems. For overdetermined and underdetermined systems, variants like Newton-Raphson (NR), Newton-Krylov, and Broyden's methods use Jacobian pseudo-inverses~\cite{akram2015newton, knoll2004jacobian, griewank2012broyden, ge2016rrt, berenson2011constrained, buss2004introduction}. However, in systems with many concatenated constraints and high dimensional configuration spaces, these methods diminish in effectiveness due to their holistic consideration of constraints, which struggles to minimize individual constraint errors. Consequently, Jacobian pseudoinverse-based approaches are inadequate for coordinated multi-robot manipulation tasks.

Compared to the prior approaches, Constellation~\cite{kaiser2012constellation} is a cyclic method. It iteratively solves for each constraint, computes the centroid of these solutions, and repeats.
While capable of addressing many constraints, its poor convergence time precludes its integration into a planner.
The nonlinear Kaczmarz method (NKZ) is an efficient cyclic solver for low residual thresholds. It is a gradient descent-based method that computes the Jacobian of a single constraint per iteration rather than the entire Jacobian matrix~\cite{wang2022nonlinear}. This cyclic approach scales better to high-dimensional configuration spaces. Compared to previously discussed methods, it improves convergence rates, making NKZ suitable for coordinated multi-robot tasks for moving structural parts together. However, NKZ has not been used in a robotics context, partially due to its inability to define constraint-specific residual thresholds. It fails to account for constraints representing different physical quantities (e.g., velocity, distance, angular, and trigonometric measurements). Our proposed constrained nonlinear Kaczmarz (cNKZ) optimizer introduces manifold-level residual thresholds, accommodating distinct nonlinear properties of each manifold (\cref{section:cnkz}).

\section{Constraints Modeled as Manifolds}\label{section:manifold_model}

As discussed in \cref{section:introduction} and \cref{section:relatedwork}, multiple robots cooperatively maneuvering a structure induces various constraints to be satisfied simultaneously. There are difficulties in representing and handling multiple constraints simultaneously in current nonlinear optimizers \cite{englert2021sampling}. To solve multiple constraints simultaneously, we develop a generalized constraint representation, which we use to address constraints such as:
\begin{itemize}
    \item Structural Constraints
    \item Task-induced Constraints
    \item Robot-specific Operational Constraints
\end{itemize}
We present the notation in \cref{section:problem_formulation} and outline our unified manifold definition in \cref{section:struct,section:task,section:robot}. It is important to note that the constraints listed above are not exhaustive. Task-induced and robot-specific operational constraints can change depending on the task requirements and robot limitations. The constraints defined are tailored to the specific nature of our tasks and the robotic hardware used in our experiments.

\subsection{Preliminaries}\label{section:problem_formulation}
We have a team of $n$ AMMs operating in collaboration to manipulate a rigid structure $S$ with a known shape and pre-defined contact points that represent specific grips (\cref{figure:problem_formulation}). We denote the state configuration of the $i^{th}$ AMM as $\boldsymbol{q}_{i} \in C=\mathbb{R}^{r}$ with $r$ degrees of freedom (DoF). This formulation presents a kinematic motion planning problem for $\barr{q} = [\boldsymbol{q}_{1}^{T},\cdots,\boldsymbol{q}_{n}^{T}]^{T}\in \mathbb{C} = C^{n}$, representing the collective state configuration (system configuration) of all $n$ AMMs having $m = n \times r$ DoF.

Let us denote an obstacle-free configuration space by ${C}_{free}$. The contact point $e_{i}$ coincides with the end-effector pose for the $i^{th}$ AMM. Hence, we can define contact configuration of the $i^{th}$ AMM by 
$\boldsymbol{p}_{i} = [f_{pos}(\boldsymbol{q}_{i}),f_{ori}(\boldsymbol{q}_{i})]$ where $f$ is the forward kinematics function.

While carrying the structure cooperatively, the robot team has to resolve all the constraints implied on them due to structural rigidity, task requirements, and robot limitations. Let us consider that there are $N$ such constraints together. These constraints are formulated as a manifold $M_{j}:=\boldsymbol{h}_{M_{j}} ( \barr{q} ): \R^{m} \to \R^{l_{j}}$, $j\in\{1,\cdots,N\}$. The sampled system configuration $\boldsymbol{\barr{q}}$ is projected on the intersection of all $N$ manifolds to generate configurations in ${C}_{satisfy}$ that satisfy all the constraints that are imposed on the end-effectors simultaneously and lie in ${C}_{free}$.

\subsection{Structural Constraints}\label{section:struct}

While carrying a structure, the motion of the robots' end-effectors is limited due to the inherent rigidity of the structure, which we refer to as \emph{Structural Constraints}. Assuming that the robots have securely grasped the structure at specified positions, these contact points will maintain constant spatial relationships. Thus, we leverage the distance and angular properties to define the structural constraints comprehensively.

\begin{definition}[Fixed Distance between Contact Points ($M_{1}$)] \label{section:man1}
Let $L_{i,j}$ be the distance between the ${i}^{th}$ and the ${j}^{th}$ contact point on the structure. For a comprehensive set, we consider all the possible distinct pairs of contact points. Hence, we obtain the manifold $StructureFixedDistance (M_{1}): \boldsymbol{h}_{M_1}(\barr{q})$ represented by:
\begin{equation}\label{eq:m1}
    \begin{aligned}
    &\boldsymbol{h}_{M_1}(\barr{q}) = [\cdots,\big|\lVert f_{pos}(\boldsymbol{q}_{i}) - f_{pos}(\boldsymbol{q}_{j}) \rVert - L_{i,j} \big|,\cdots]^{T} \\
    &\forall \; \{i,j\} \in \{\{i,j\} \;|\; i \neq j \;\text{and}\; i,j \in  \{1,\cdots,n\}\}. 
    \end{aligned}\hspace{-10pt}
\end{equation}
\end{definition}

\begin{definition}[Fixed Angles between Contact Points ($M_{2}$)]\label{section:man2}

Let us consider a set of three distinct contact points on the structure $S$ defined by ${H}_{pairs} = \{\{i,j,k\} \;|\; i \neq j \neq k \;\& \; i,j,k \in \{1,\cdots,n\}\}$. Furthermore, we define a vector $\mathbf{v}_{(a,b)}$ for any pair of contact points $(a,b)$ on the structure. In the case of contact points ${i,j,k}$, the angle between $\mathbf{v}_{(i,j)}$ and $\mathbf{v}_{(j,k)}$, $\angle\mathbf{v}_{(i,j)}\mathbf{v}_{(j,k)}$, remains constant regardless of the robot states $\{\boldsymbol{q}_{i},\boldsymbol{q}_{j},\boldsymbol{q}_{k}\}$. However, to ensure successful motion, the fixed angle between these contact points must align with the angle calculated from the end-effector positions, obtained through the robot's forward kinematics. We establish $\mathbf{V}_{a,b}= f_{pos}(\boldsymbol{q}_{a}) - f_{pos}(\boldsymbol{q}_{b})$ as the vector between the contact points $a$ and $b$ of the AMMs.

Mathematically, the angle between two vectors can be computed using the dot product of these vectors. We leverage these mathematical principles to derive $StructureFixedAngle (M_{2}): \boldsymbol{h}_{M_2}(\barr{q})$~(\cref{eq:m2}):
\begin{equation}\label{eq:m2}
    \begin{aligned}
    \boldsymbol{h}_{M_2}(\barr{q}) =  [\cdots,\mathbf{V}_{i,j} \cdot \mathbf{V}_{j,k} - \cos(&\angle\mathbf{v}_{(i,j)}\mathbf{v}_{(j,k)}),\cdots]^{T}\\
    &\forall \; \{i,j,k\} \in {H}_{pairs}.
    \end{aligned}\hspace{-10pt}
\end{equation}

Furthermore, we utilize cross product to check vector collinearity. Note that $M_{2}$ angular constraints apply only to teams of three or more robots, as a minimum of three contact points is required to define an angle.
\end{definition}

\subsection{Task-induced Constraints}\label{section:task}
Based on the nature of the task (e.g., carrying a tank filled with water), constraints can be applied to facilitate robot and object motions. We refer to this set of task-specific constraints as \emph{Task-induced constraints}. We establish a constraint to fix the end-effectors' relative orientation \cref{section:man4} to adhere to the end-effectors constraints for real-world experiments~(\cref{figure:problem_formulation}). While not explicitly addressed in this work, additional task-specific constraints (e.g., restricting structure orientation or movement to a fixed plane) can be defined using similar methods.

\begin{definition}[Restricted Orientation between End-effectors and their Respective Contact Points ($M_{3}$)]\label{section:man4}

Due to the shape of the end-effector, we introduce a constraint to maintain orthogonality between the end-effector and its respective contact points. Given collinear contact points, we restrict the end-effector orientation to be orthogonal to a vector formed by any two contact points. Hence, we define the manifold $TaskFixedOrient (M_{3}): \boldsymbol{h}_{M_3}(\barr{q})$ as:
\begin{equation}\label{eq:m4}
    \begin{aligned}
        \boldsymbol{h}_{M_3}(\barr{q}) = [\cdots,\big(f_{pos}(\boldsymbol{q}_{i}) -  f_{pos}&(\boldsymbol{q}_{j})\big) \cdot f_{ori}(\boldsymbol{q}_{k}),\cdots]^{T},\\
        &\forall \; \{i,j,k\} \in {H}_{pairs}.
    \end{aligned}\hspace{-10pt}
\end{equation}

\end{definition}

\begin{definition}[Maintaining Structure on the Same Plane ($M_{4}$)]\label{section:man5}
An additional task constraint on the structure's orientation is applied to facilitate a balanced load amongst the AMMs. For example, we constrained the structure to be horizontal (in the X-Y plane) at all times. In this situation, the contact points of individual AMMs satisfy the same plane, hence giving $TaskSamePlane (M_{4}): \boldsymbol{h}_{M_4}(\barr{q})$ as:

\begin{equation}\label{eq:m5}
    \begin{aligned}
        \boldsymbol{h}_{M_4}(\barr{q}) = 
        [\cdots, P(f_{pos}(\boldsymbol{q}_{i})) - P&(f_{pos}(\boldsymbol{q}_{j})),\cdots]^{T},\\
        &\forall \; \{i,j,k\} \in {H}_{pairs}.
    \end{aligned}
\end{equation}

$P$ represents a function that imposes this constraint, e.g., to maintain horizontal movement, $P(\boldsymbol{p}_{i}) = \boldsymbol{p}_{i,z}$, where  $\boldsymbol{p}_{i,z}$ represents the z-coordinate of $\boldsymbol{p}_{i}$.

\end{definition}

\subsection{Robot-specific Operational Constraints}\label{section:robot}

\begin{definition}[Differential-drive-based Nonholonomic Robot ($M_5$)]\label{section:man6}
One of the most common nonholonomic constraints is the differential drive mechanism, which depends on the robot's velocity. We consider a nonholonomic differential-drive-based ground robot with configuration $q_{i,x}, q_{i,y}, q_{i, yaw}$ and control input velocities $\boldsymbol{u}(t) = [u_{i,x}, u_{i,y}]$ for the $i^{th}$ robot. Then, we can represent the manifold $RobotDiffDrive (M_5)$ as:
\begin{equation}\label{eq:m6}
    \begin{aligned}
        \boldsymbol{h}_{M_5}(\barr{q}) = [\cdots, u_{i,x} \sin({q_{i,yaw}})-&u_{i,y}\cos({q_{i,yaw}}),\cdots]^{T}\\
        &\forall i \in \{1,\cdots,n\}\}.
    \end{aligned}\hspace{-10pt}
\end{equation}

\end{definition}

\section{Projection on Intersection of Manifolds}\label{section:project}

We propose the constrained nonlinear Kaczmarz (cNKZ) algorithm (\cref{algo:cnkzplan}), a nonlinear projection method that generates configurations satisfying all constraints simultaneously in \cref{section:cnkz}, followed by a centralized cooperative motion planner in\cref{section:planning}. While we integrate cNKZ into an RRT for cooperative and coordinated manipulation in clutter, its applicability extends beyond, demonstrating potential for solving multiple constraints in high-dimensional spaces across various domains.

\subsection{Constrained Nonlinear Kaczmarz Algorithm (cNKZ)}\label{section:cnkz}
As outlined in \cref{section:relatedwork}, the global residual threshold approach of nonlinear Kaczmarz (NKZ) for constraint systems proves inadequate for our system's diverse manifolds, which have varying sensitivity levels. For example, $M_1$ residual denotes distance error, while $M_2$ represent angular errors. We address this limitation by implementing weighted manifold sets, enabling fine-tuning for sensitivity adjustments for specific manifolds. Moreover, cNKZ allows the definition of independent residual thresholds for each manifold, offering more precise accuracy control and accommodating the distinct nonlinear properties arising from the manifolds' functional dependencies (e.g., the trigonometric $\cos$ function in $M_2$). 

Let us refer to the manifold-specific residual threshold as the \emph{manifold threshold}. Note that each manifold has a set of constraints, each corresponding to an intrinsic dimension. Every iteration selects an individual constraint and evaluate it against the predetermined \emph{manifold threshold}. If unsatisfied, the corresponding Jacobian is computed to generate an updated configuration. This single-constraint update approach reduces computational complexity and accelerates convergence in the projection process.~(\cref{fig:residuals}). The process continues till all constraints are satisfied with the required individual accuracy~(\cref{algo:cnkzplan}).

\algnewcommand\algorithmicreturn{\textbf{return}}
\algnewcommand\RETURN{\State \algorithmicreturn}%
\begin{algorithm}
\begin{algorithmic}[1]
\STATE {Initialize: $\boldsymbol{q}^{(0)} = \barr{q}_{0},\; step = 0,\; \boldsymbol{r} = - \boldsymbol{F}(\boldsymbol{q}^{(0)})$}
\STATE {$\boldsymbol{r_{\scriptscriptstyle{status}}}=\boldsymbol{|r|} \leq \boldsymbol{\barr{\epsilon}}$}
\COMMENT{While any manifold threshold is unsatisfied}
\WHILE{$\min{(\boldsymbol{r_{\scriptscriptstyle{status}}})} \neq True \; \& \; step \leqslant \gamma$}
    \STATE {$i = step \pmod{l}$}
    \COMMENT{Update if constraint $i$ is unsatisfied}
    \IF {$\boldsymbol{r_{\scriptscriptstyle{status}}}[i] \neq True $}
        \COMMENT{Compute constraint specific Jacobian}
        \STATE {$\boldsymbol{g}_{i} = \boldsymbol{J}_{i}(\boldsymbol{q}^{(step)})$}
        \COMMENT{Project on hyperplane of i$^{th}$ constraint}
        \STATE {$\boldsymbol{q}^{(step+1)} \xleftarrow{} \boldsymbol{q}^{(step)} - \frac{\boldsymbol{r}(i)}{\|\boldsymbol{g}_{i}\|^{2}}\cdot\boldsymbol{g}_{i}$}
        \STATE {$\boldsymbol{r}_{new} = -\boldsymbol{F}(\boldsymbol{q}^{(step+1)})$} \label{algo:residual}
    \ENDIF
    \COMMENT{Update $\barr{q}_{proj}$}
    \IF {$\|\boldsymbol{r}_{new}\| < \|\boldsymbol{r}\|$}
        \STATE {$\barr{q}_{proj} \xleftarrow{} \boldsymbol{q}^{(step + 1)}$}
        \STATE {$\boldsymbol{r} = \boldsymbol{r}_{new}$}
    \ENDIF
    \STATE{$step = step + 1$}
\ENDWHILE
\RETURN $~\barr{q}_{proj}$ 
\end{algorithmic}
    \caption{PROJECT\_cNKZ($\boldsymbol{F}, \barr{q}_{0}, \gamma, \boldsymbol{\barr{\epsilon}}$)}
    \label{algo:cnkzplan}
\end{algorithm}

We can define a system of nonlinear equations $\mathbf{F}(\barr{q}) \leq \boldsymbol{\barr{\epsilon}}$ by concatenating all the constraints and their respective manifold thresholds to obtain $\boldsymbol{\barr{\epsilon}}$ (\cref{eq:f_x}). Here, $\mathbf{F}: \R^{m} \xrightarrow{} \R^{l}$ is a nonlinear vector-valued function, and $\barr{q} \in \R^{n}$ is the valid configuration in $\mathbb{C}$. Here, $m$ is the number of system-level DoF, ${l}$ is the total number of nonlinear equations, and $n$ is the number of robots:
\begin{equation}\label{eq:f_x}
    \begin{aligned}
        \boldsymbol{F}(\barr{q}) = [\cdots,(\boldsymbol{h}_{M_{i}}(\barr{q}))^{T},\cdots]^{T} & \quad \forall \: i \in \{1,\cdots,N\},
    \end{aligned}\hspace{-10pt}
\end{equation}
where $N$ is the number of manifolds. Our target is to find $\barr{q}^{*}$ such that $\boldsymbol{F}(\barr{q}^{*}) \leq \boldsymbol{\barr{\epsilon}}$. In every iteration, cNKZ obtains the residual corresponding to a single hyperplane to give an updated configuration $\barr{q}^{'}$:
\begin{equation}\label{eq:kacz_approx}
    \begin{aligned}
    \barr{q}^{'} = \barr{q} - \frac{\boldsymbol{F}_{i}(\barr{q})}{\|\boldsymbol{J}_{i}(\barr{q})\|^{2}}\boldsymbol{J}_{i}(\barr{q}),
    \end{aligned}
\end{equation}
where $\boldsymbol{J}_{i}$ is the $i^{th}$ row of the jacobian ($\boldsymbol{J}$) of $\boldsymbol{F}$.

\subsection{Multi-Constrained Motion Planning Algorithm} \label{section:planning}

To generate an executable trajectory, we extend the RRT algorithm~\cite{lavalle1998rapidly}, capitalizing on its flexibility and scalability to higher-dimensional configuration spaces~\cite{zhang2021task}. We adapt the conventional steer-to-extend functionality, integrating the cNKZ projection method (illustrated in~\cref{algo:cnkzplan}) to produce control inputs for constrained trajectories applicable to multi-robot teams~\cite{englert2021sampling, berenson2011constrained}. Additionally, we employ a rejection strategy to ensure compliance with joint limits and perform collision checking, thereby maintaining feasibility.

\section{Experiments and Results}\label{sec:sim}
\begin{table*}[]
    \centering
    \caption{Comparative analysis of projection algorithms for constraint representing manifolds. The table presents performance metrics for various constraint-imposing manifolds and their combinations, applied to a team of three robots (18 DoF) transporting a linear structure. The leftmost column specifies the constraint manifolds and the mapping from system configuration space to intrinsic dimensionality of constraints (in parentheses). Performance is evaluated using average computation time and success rate for the proposed cNKZ projection algorithm, compared against existing methods: NR, CIM, and NKZ. Note that '--' represents all failures.}
    \scriptsize
    \renewcommand{\arraystretch}{1.3}
    \begin{tabularx}{\textwidth}{c|c|cccc|YYYY}
Manifolds	&	Residual	&	\multicolumn{4}{c|}{Success Rate (\%)}	&	\multicolumn{4}{c}{Average Computation Time (Seconds)}	\\ 
(${{\mathbb{R}^{\scalebox{.8}{m}}\rightarrow \mathbb{R}^{\scalebox{.8}{l}}}}$)	&	Threshold	&	\cellcolor{white!0}NR	&	\cellcolor{lgray}CIM	&	\cellcolor{white!0}NKZ	&	\cellcolor{lgray}cNKZ	&	\cellcolor{white!0}NR	&	\cellcolor{lgray}CIM	&	\cellcolor{white!0}NKZ	&	\cellcolor{lgray}cNKZ	 \\ 	 \hhline{==========} 
$M_1$, $M_3$ & 	0.5 & 	\cellcolor{white!0}0.0 & 	\cellcolor{lgray}5.0 & 	\cellcolor{white!0}$\mathbf{100.0}$ & 	\cellcolor{lgray}$\mathbf{100.0}$ & 	\cellcolor{white!0} -- & 	\cellcolor{lgray}$0.392{\scriptscriptstyle\pm} 0.18$ & 	\cellcolor{white!0}$0.019{\scriptscriptstyle\pm} 0.01$ & 	\cellcolor{lgray}$\mathbf{0.016{\scriptscriptstyle\pm} 0.0}$ \\ 
(${{\mathbb{R}^{\scalebox{.8}{18}}\rightarrow \mathbb{R}^{\scalebox{.8}{6}}}}$) & 	6.0 & 	\cellcolor{white!0}0.0 & 	\cellcolor{lgray}0.0 & 	\cellcolor{white!0}$\mathbf{100.0}$ & 	\cellcolor{lgray}$\mathbf{100.0}$ & 	\cellcolor{white!0} -- & 	\cellcolor{lgray} -- & 	\cellcolor{white!0}$0.067{\scriptscriptstyle\pm} 0.02$ & 	\cellcolor{lgray}$\mathbf{0.042{\scriptscriptstyle\pm} 0.01}$ \\ \hline
 $M_3$, $M_4$& 	0.5 & 	\cellcolor{white!0}0.0 & 	\cellcolor{lgray}6.5 & 	\cellcolor{white!0}$\mathbf{100.0}$ & 	\cellcolor{lgray}99.5 & 	\cellcolor{white!0} -- & 	\cellcolor{lgray}$1.978{\scriptscriptstyle\pm} 1.02$ & 	\cellcolor{white!0}$0.15{\scriptscriptstyle\pm} 0.08$ & 	\cellcolor{lgray}$\mathbf{0.1{\scriptscriptstyle\pm} 0.03}$ \\ 
(${{\mathbb{R}^{\scalebox{.8}{18}}\rightarrow \mathbb{R}^{\scalebox{.8}{8}}}}$) & 	2.0 & 	\cellcolor{white!0}1.0 & 	\cellcolor{lgray}10.5 & 	\cellcolor{white!0}$\mathbf{100.0}$ & 	\cellcolor{lgray}$\mathbf{100.0}$ & 	\cellcolor{white!0}$0.118{\scriptscriptstyle\pm} 0.1$ & 	\cellcolor{lgray}$1.578{\scriptscriptstyle\pm} 1.01$ & 	\cellcolor{white!0}$0.092{\scriptscriptstyle\pm} 0.03$ & 	\cellcolor{lgray}$\mathbf{0.054{\scriptscriptstyle\pm} 0.02}$ \\ \hline
 $M_1$, $M_2$, $M_3$& 	0.5 & 	\cellcolor{white!0}0.0 & 	\cellcolor{lgray}0.0 & 	\cellcolor{white!0}$\mathbf{100.0}$ & 	\cellcolor{lgray}$\mathbf{100.0}$ & 	\cellcolor{white!0} -- & 	\cellcolor{lgray} -- & 	\cellcolor{white!0}$0.356{\scriptscriptstyle\pm} 0.17$ & 	\cellcolor{lgray}$\mathbf{0.16{\scriptscriptstyle\pm} 0.04}$ \\ 
(${{\mathbb{R}^{\scalebox{.8}{18}}\rightarrow \mathbb{R}^{\scalebox{.8}{12}}}}$) & 	2.0 & 	\cellcolor{white!0}0.0 & 	\cellcolor{lgray}0.0 & 	\cellcolor{white!0}$\mathbf{100.0}$ & 	\cellcolor{lgray}$\mathbf{100.0}$ & 	\cellcolor{white!0} -- & 	\cellcolor{lgray} -- & 	\cellcolor{white!0}$0.171{\scriptscriptstyle\pm} 0.04$ & 	\cellcolor{lgray}$\mathbf{0.091{\scriptscriptstyle\pm} 0.02}$ \\ \hline
 $M_1$, $M_2$, $M_3$, $M_4$& 	0.5 & 	\cellcolor{white!0}0.0 & 	\cellcolor{lgray}0.0 & 	\cellcolor{white!0}$\mathbf{100.0}$ & 	\cellcolor{lgray}$\mathbf{100.0}$ & 	\cellcolor{white!0} -- & 	\cellcolor{lgray} -- & 	\cellcolor{white!0}$1.02{\scriptscriptstyle\pm} 0.52$ & 	\cellcolor{lgray}$\mathbf{0.392{\scriptscriptstyle\pm} 0.11}$ \\ 
(${{\mathbb{R}^{\scalebox{.8}{18}}\rightarrow \mathbb{R}^{\scalebox{.8}{14}}}}$) & 	6.0 & 	\cellcolor{white!0}0.0 & 	\cellcolor{lgray}0.0 & 	\cellcolor{white!0}$\mathbf{100.0}$ & 	\cellcolor{lgray}$\mathbf{100.0}$ & 	\cellcolor{white!0} -- & 	\cellcolor{lgray} -- & 	\cellcolor{white!0}$0.168{\scriptscriptstyle\pm} 0.06$ & 	\cellcolor{lgray}$\mathbf{0.094{\scriptscriptstyle\pm} 0.02}$ \\ \hline
\end{tabularx}
\label{table:projection}
\end{table*}

In this section, we analyze the performance of cNKZ, our proposed projection algorithm, on optimizing for system-level state configurations over various combinations of manifold and robots and compare its performance against baseline nonlinear optimization approaches. We also analyze the success rate and computational time required for generating motion plans and associated projections for transporting straight, T-shaped, and I-shaped structures using up to six AMMs (up to 36 DoF). We compare transporting different structures based on the following criteria:
\begin{itemize}
    \item \textbf{Success Rate:} Average number of successful projections on the intersection of manifolds and successful path generation rate over multiple trials.
    \item \textbf{Computation Time:} Average computational time required to generate successful paths and projections.
\end{itemize}

To assess the performance in 3D environments, we simulated teams of AMMs where the base robot had dimensions of $0.4~m~\times~0.4~m~\times~0.4~m$ and can maneuver in 3D space with four DoF ($x,\;y,\;z,\;yaw$). The mobile manipulator was equipped with a 2-DoF robotic arm with link lengths of $0.2~m$ and $0.1~m$ and operated in the Y-Z plane of the base robot's fixed body frame. All experiments were performed on a laptop with an Intel Core i7-12700H CPU.
\begin{figure}[t]
    \centering
    \includegraphics[width=8.5cm, trim={0cm 0cm 0cm 0cm}, clip]{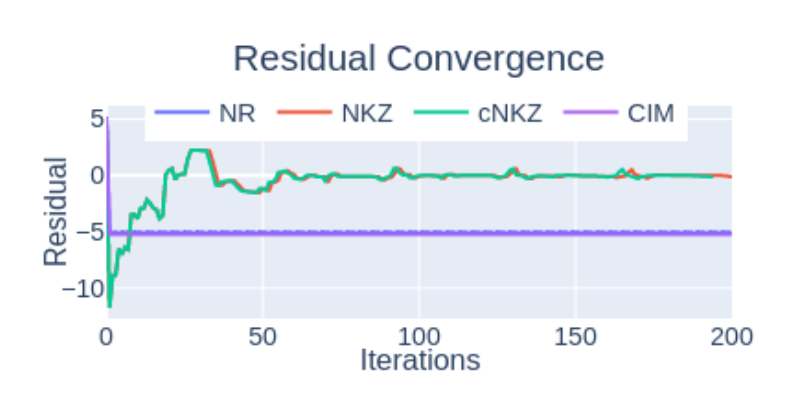}
    \caption{An example illustrative behavior of residual progression over iterations for a constituent constraint (from 38 constraints simultaneously solved for $I$ shaped structure using $5$ robot team (30 DoF)). cNKZ converges the fastest, followed by NKZ, whereas NR gets stuck in a local minima loop.}
    \label{fig:residuals}
\end{figure}
\subsection{Different Projection Techniques}
We evaluate the precision of constrained nonlinear Kaczmarz (cNKZ) through comparative analysis with Newton-Raphson (NR), nonlinear Cimmino's (CIM), and nonlinear Kaczmarz (NKZ) methods ~\cite{akram2015newton, nonlinearcimmino, wang2022nonlinear}. We compare the success rate and projection time to assess cNKZ's efficacy for performing efficient projection.

We randomly sample configurations for three AMMs operating within predefined environments in our experimental setup. Subsequently, NR, CIM, NKZ, and cNKZ project each sample onto the intersection of manifolds. We considered structural ($M_{1}\;and\;M_{2}$) and task-enforced constraints ($M_{3}\; and \; M_{4}$) for three AMMs tasked with transporting a linear structure spanning $1~m$ with fixed contact points $L=0.5~m$ interval. We conducted a comparison of various projection techniques across 200 simulated trials.

\cref{table:projection} demonstrates cNKZ vastly outperforming other projection algorithms. NKZ and cNKZ achieved high success rates, while NR and CIM struggled. cNKZ halved projection time versus NKZ as manifolds increased. For an I-shaped structure constraint, NKZ and cNKZ both converged, although cNKZ is faster (\cref{fig:residuals}). NR is stuck in loop due to pseudo-Jacobian dependence. While applicable to various equation system ranks—full, overdetermined, or underdetermined—cNKZ excelled in success rate and computation time.

\begin{table}[t]
\centering
\renewcommand{\arraystretch}{1.3}
\caption{Comparison of structural constraint residuals: Constraints as structural manifolds (\cref{section:struct}) results to  $100\times$ lower residuals compared to TSR.}
\scriptsize
\begin{tabular}{c|c|c}
Structural Constraint & TSR $+$ NR(Residual)              & Ours $+$ cNKZ (Residual)                   \\ \hhline{===}
$M_{1}$ & $0.117~\pm~0.074$ & $\mathbf{0.0054~\pm~0.0043}$ \\
$M_{2}$ & $0.397~\pm~0.306$ & $\mathbf{0.0003~\pm~0.0004}$ \\ \hline
\end{tabular}
\label{table:repcompare}
\end{table}

\begin{table*}[ht]
    \renewcommand{\arraystretch}{1.3}
    \scriptsize
    \caption{Performance analysis of centralized constrained motion planning for coordinated multi-robot tasks using the proposed cNKZ projection algorithm integrated with RRT. The left panel illustrates the structural geometry, with colored dots representing contact points for the robot end-effectors, and outlines the mapping from system configuration space to intrinsic dimensionality of constraints ($\R^{m} \rightarrow \R^{l}$). Success rate percentages include successful plans across environments of varying complexity (determined by cluttered space volume) and the average percentage of successful projections per plan. Computation time statistics present mean ± standard deviation (in minutes) for generating a successful plan and producing a successful projection per plan (in seconds).}
    \begin{tabularx}{\textwidth}{I|l|YYYY|cccc}
\multicoltext{$\R^{m} \xrightarrow{} \R^{l}$}{Constraints}{(Structure\_\#Robots)}	&		&	\multicolumn{4}{c|}{Success Rate (\%)}	&	\multicolumn{4}{c}{Average Computation Time}	\\ 
	&	\multicolumn{1}{c|}{Environmental}	&	\multicolumn{2}{c}{Planning}	&	\multicolumn{2}{c|}{Projection}	&	\multicolumn{2}{c}{Planning (minutes)}	&	\multicolumn{2}{c}{Per Projection (seconds)}	 \\ 
	&	\multicolumn{1}{c|}{Complexity}	&	\cellcolor{white!0}NKZ	&	\cellcolor{lgray}cNKZ	&	\cellcolor{white!0}NKZ	&	\cellcolor{lgray}cNKZ	&	\cellcolor{white!0}NKZ	&	\cellcolor{lgray}cNKZ	&	\cellcolor{white!0}NKZ	&	\cellcolor{lgray}cNKZ	 \\ 	 \hhline{==========} 
\tblimgtext{T_3.png}{${{\mathbb{R}^{\scalebox{.8}{18}}\rightarrow \mathbb{R}^{\scalebox{.8}{12}}}}$}{T\_3} & 	Low & 	\cellcolor{white!0}76.0 & 	\cellcolor{lgray}$\mathbf{90.0}$ & 	\cellcolor{white!0}$\mathbf{100.0}$ & 	\cellcolor{lgray}$\mathbf{100.0}$ & 	\cellcolor{white!0}$2.04{\scriptscriptstyle\pm} 0.6$ & 	\cellcolor{lgray}$\mathbf{1.11{\scriptscriptstyle\pm} 0.43}$ & 	\cellcolor{white!0}$1.22{\scriptscriptstyle\pm} 0.36$ & 	\cellcolor{lgray}$\mathbf{0.59{\scriptscriptstyle\pm} 0.16}$ \\ 
 & 	Medium & 	\cellcolor{white!0}60.0 & 	\cellcolor{lgray}$\mathbf{74.0}$ & 	\cellcolor{white!0}$\mathbf{100.0}$ & 	\cellcolor{lgray}$\mathbf{100.0}$ & 	\cellcolor{white!0}$1.85{\scriptscriptstyle\pm} 0.58$ & 	\cellcolor{lgray}$\mathbf{1.04{\scriptscriptstyle\pm} 0.4}$ & 	\cellcolor{white!0}$1.1{\scriptscriptstyle\pm} 0.35$ & 	\cellcolor{lgray}$\mathbf{0.54{\scriptscriptstyle\pm} 0.15}$ \\ 
 & 	Hard & 	\cellcolor{white!0}62.0 & 	\cellcolor{lgray}$\mathbf{76.0}$ & 	\cellcolor{white!0}$\mathbf{100.0}$ & 	\cellcolor{lgray}$\mathbf{100.0}$ & 	\cellcolor{white!0}$1.63{\scriptscriptstyle\pm} 0.46$ & 	\cellcolor{lgray}$\mathbf{1.17{\scriptscriptstyle\pm} 0.5}$ & 	\cellcolor{white!0}$0.97{\scriptscriptstyle\pm} 0.28$ & 	\cellcolor{lgray}$\mathbf{0.56{\scriptscriptstyle\pm} 0.17}$ \\ \hline
\tblimgtext{S_3.png}{${{\mathbb{R}^{\scalebox{.8}{18}}\rightarrow \mathbb{R}^{\scalebox{.8}{14}}}}$}{S\_3} & 	Low & 	\cellcolor{white!0}80.0 & 	\cellcolor{lgray}$\mathbf{92.0}$ & 	\cellcolor{white!0}91.65 & 	\cellcolor{lgray}$\mathbf{97.13}$ & 	\cellcolor{white!0}$3.72{\scriptscriptstyle\pm} 1.87$ & 	\cellcolor{lgray}$\mathbf{1.83{\scriptscriptstyle\pm} 0.49}$ & 	\cellcolor{white!0}$2.23{\scriptscriptstyle\pm} 1.12$ & 	\cellcolor{lgray}$\mathbf{1.05{\scriptscriptstyle\pm} 0.37}$ \\ 
 & 	Medium & 	\cellcolor{white!0}66.0 & 	\cellcolor{lgray}$\mathbf{76.0}$ & 	\cellcolor{white!0}91.94 & 	\cellcolor{lgray}$\mathbf{97.05}$ & 	\cellcolor{white!0}$3.8{\scriptscriptstyle\pm} 1.64$ & 	\cellcolor{lgray}$\mathbf{1.73{\scriptscriptstyle\pm} 0.54}$ & 	\cellcolor{white!0}$2.28{\scriptscriptstyle\pm} 0.98$ & 	\cellcolor{lgray}$\mathbf{1.0{\scriptscriptstyle\pm} 0.38}$ \\ 
 & 	Hard & 	\cellcolor{white!0}78.0 & 	\cellcolor{lgray}$\mathbf{86.0}$ & 	\cellcolor{white!0}92.38 & 	\cellcolor{lgray}$\mathbf{97.07}$ & 	\cellcolor{white!0}$4.36{\scriptscriptstyle\pm} 1.54$ & 	\cellcolor{lgray}$\mathbf{1.78{\scriptscriptstyle\pm} 0.49}$ & 	\cellcolor{white!0}$2.62{\scriptscriptstyle\pm} 0.92$ & 	\cellcolor{lgray}$\mathbf{1.04{\scriptscriptstyle\pm} 0.34}$ \\ \hline
\tblimgtext{S_4.png}{${{\mathbb{R}^{\scalebox{.8}{24}}\rightarrow \mathbb{R}^{\scalebox{.8}{30}}}}$}{S\_4} & 	Low & 	\cellcolor{white!0}84.0 & 	\cellcolor{lgray}$\mathbf{86.0}$ & 	\cellcolor{white!0}$\mathbf{100.0}$ & 	\cellcolor{lgray}$\mathbf{100.0}$ & 	\cellcolor{white!0}$13.86{\scriptscriptstyle\pm} 3.27$ & 	\cellcolor{lgray}$\mathbf{2.82{\scriptscriptstyle\pm} 1.15}$ & 	\cellcolor{white!0}$8.31{\scriptscriptstyle\pm} 1.96$ & 	\cellcolor{lgray}$\mathbf{1.58{\scriptscriptstyle\pm} 0.56}$ \\ 
 & 	Medium & 	\cellcolor{white!0}$\mathbf{82.0}$ & 	\cellcolor{lgray}$\mathbf{82.0}$ & 	\cellcolor{white!0}$\mathbf{100.0}$ & 	\cellcolor{lgray}$\mathbf{100.0}$ & 	\cellcolor{white!0}$14.19{\scriptscriptstyle\pm} 3.67$ & 	\cellcolor{lgray}$\mathbf{3.12{\scriptscriptstyle\pm} 1.21}$ & 	\cellcolor{white!0}$8.51{\scriptscriptstyle\pm} 2.2$ & 	\cellcolor{lgray}$\mathbf{1.59{\scriptscriptstyle\pm} 0.44}$ \\ 
 & 	Hard & 	\cellcolor{white!0}82.0 & 	\cellcolor{lgray}$\mathbf{84.0}$ & 	\cellcolor{white!0}$\mathbf{100.0}$ & 	\cellcolor{lgray}$\mathbf{100.0}$ & 	\cellcolor{white!0}$12.44{\scriptscriptstyle\pm} 2.95$ & 	\cellcolor{lgray}$\mathbf{3.0{\scriptscriptstyle\pm} 1.17}$ & 	\cellcolor{white!0}$7.46{\scriptscriptstyle\pm} 1.77$ & 	\cellcolor{lgray}$\mathbf{1.56{\scriptscriptstyle\pm} 0.53}$ \\ \hline
\tblimgtext{I_5.png}{${{\mathbb{R}^{\scalebox{.8}{30}}\rightarrow \mathbb{R}^{\scalebox{.8}{37}}}}$}{I\_5} & 	Low & 	\cellcolor{white!0}66.0 & 	\cellcolor{lgray}$\mathbf{78.0}$ & 	\cellcolor{white!0}73.61 & 	\cellcolor{lgray}$\mathbf{99.4}$ & 	\cellcolor{white!0}$120.74{\scriptscriptstyle\pm} 31.4$ & 	\cellcolor{lgray}$\mathbf{14.03{\scriptscriptstyle\pm} 3.46}$ & 	\cellcolor{white!0}$60.36{\scriptscriptstyle\pm} 15.7$ & 	\cellcolor{lgray}$\mathbf{7.01{\scriptscriptstyle\pm} 1.73}$ \\ 
 & 	Medium & 	\cellcolor{white!0}62.0 & 	\cellcolor{lgray}$\mathbf{72.0}$ & 	\cellcolor{white!0}74.19 & 	\cellcolor{lgray}$\mathbf{99.75}$ & 	\cellcolor{white!0}$111.39{\scriptscriptstyle\pm} 25.49$ & 	\cellcolor{lgray}$\mathbf{14.01{\scriptscriptstyle\pm} 3.99}$ & 	\cellcolor{white!0}$54.38{\scriptscriptstyle\pm} 13.52$ & 	\cellcolor{lgray}$\mathbf{7.0{\scriptscriptstyle\pm} 2.0}$ \\ 
 & 	Hard & 	\cellcolor{white!0}50.0 & 	\cellcolor{lgray}$\mathbf{62.0}$ & 	\cellcolor{white!0}76.63 & 	\cellcolor{lgray}$\mathbf{99.65}$ & 	\cellcolor{white!0}$93.03{\scriptscriptstyle\pm} 33.23$ & 	\cellcolor{lgray}$\mathbf{12.46{\scriptscriptstyle\pm} 3.75}$ & 	\cellcolor{white!0}$46.51{\scriptscriptstyle\pm} 16.61$ & 	\cellcolor{lgray}$\mathbf{6.23{\scriptscriptstyle\pm} 1.87}$ \\ \hline
\tblimgtext{S_5.png}{${{\mathbb{R}^{\scalebox{.8}{30}}\rightarrow \mathbb{R}^{\scalebox{.8}{52}}}}$}{S\_5} & 	Low & 	\cellcolor{white!0}$\mathbf{72.0}$ & 	\cellcolor{lgray}68.0 & 	\cellcolor{white!0}93.22 & 	\cellcolor{lgray}$\mathbf{99.56}$ & 	\cellcolor{white!0}$82.73{\scriptscriptstyle\pm} 6.38$ & 	\cellcolor{lgray}$\mathbf{8.94{\scriptscriptstyle\pm} 2.42}$ & 	\cellcolor{white!0}$49.63{\scriptscriptstyle\pm} 3.83$ & 	\cellcolor{lgray}$\mathbf{5.36{\scriptscriptstyle\pm} 1.45}$ \\ 
 & 	Medium & 	\cellcolor{white!0}46.0 & 	\cellcolor{lgray}$\mathbf{50.0}$ & 	\cellcolor{white!0}94.83 & 	\cellcolor{lgray}$\mathbf{99.12}$ & 	\cellcolor{white!0}$76.71{\scriptscriptstyle\pm} 6.61$ & 	\cellcolor{lgray}$\mathbf{8.82{\scriptscriptstyle\pm} 2.11}$ & 	\cellcolor{white!0}$46.02{\scriptscriptstyle\pm} 3.96$ & 	\cellcolor{lgray}$\mathbf{5.29{\scriptscriptstyle\pm} 1.26}$ \\ 
 & 	Hard & 	\cellcolor{white!0}28.0 & 	\cellcolor{lgray}$\mathbf{36.0}$ & 	\cellcolor{white!0}94.5 & 	\cellcolor{lgray}$\mathbf{99.06}$ & 	\cellcolor{white!0}$76.19{\scriptscriptstyle\pm} 7.84$ & 	\cellcolor{lgray}$\mathbf{6.96{\scriptscriptstyle\pm} 2.06}$ & 	\cellcolor{white!0}$45.7{\scriptscriptstyle\pm} 4.7$ & 	\cellcolor{lgray}$\mathbf{4.17{\scriptscriptstyle\pm} 1.23}$ \\ \hline
\tblimgtext{S_6.png}{${{\mathbb{R}^{\scalebox{.8}{36}}\rightarrow \mathbb{R}^{\scalebox{.8}{80}}}}$}{S\_6} & 	Low & 	\cellcolor{white!0}66.0 & 	\cellcolor{lgray}$\mathbf{68.0}$ & 	\cellcolor{white!0}60.42 & 	\cellcolor{lgray}$\mathbf{87.92}$ & 	\cellcolor{white!0}$101.03{\scriptscriptstyle\pm} 18.32$ & 	\cellcolor{lgray}$\mathbf{19.36{\scriptscriptstyle\pm} 5.71}$ & 	\cellcolor{white!0}$60.61{\scriptscriptstyle\pm} 10.99$ & 	\cellcolor{lgray}$\mathbf{10.84{\scriptscriptstyle\pm} 2.83}$ \\ 
 & 	Medium & 	\cellcolor{white!0}52.0 & 	\cellcolor{lgray}$\mathbf{60.0}$ & 	\cellcolor{white!0}62.96 & 	\cellcolor{lgray}$\mathbf{87.58}$ & 	\cellcolor{white!0}$100.23{\scriptscriptstyle\pm} 18.24$ & 	\cellcolor{lgray}$\mathbf{19.57{\scriptscriptstyle\pm} 8.03}$ & 	\cellcolor{white!0}$60.13{\scriptscriptstyle\pm} 10.94$ & 	\cellcolor{lgray}$\mathbf{10.28{\scriptscriptstyle\pm} 2.78}$ \\ 
 & 	Hard & 	\cellcolor{white!0}38.0 & 	\cellcolor{lgray}$\mathbf{56.0}$ & 	\cellcolor{white!0}61.05 & 	\cellcolor{lgray}$\mathbf{87.86}$ & 	\cellcolor{white!0}$93.19{\scriptscriptstyle\pm} 15.2$ & 	\cellcolor{lgray}$\mathbf{20.52{\scriptscriptstyle\pm} 6.4}$ & 	\cellcolor{white!0}$55.91{\scriptscriptstyle\pm} 9.12$ & 	\cellcolor{lgray}$\mathbf{10.81{\scriptscriptstyle\pm} 2.92}$ \\ \hline
\end{tabularx}
    \label{tab:planning}
\end{table*}

\subsection{Different Constraint Representation Techniques}

In cooperative motion planning, there is an absence of an algorithm that can resolve multiple constraints concurrently. For fair comparison, we focus on structural constraints, comparing against task space regions (TSR) used in multi-robot coordination~\cite{zhang2021task}. We evaluate robot configuration quality by comparing structural constraint residuals using TSR with NR method and Family of manifolds (structural manifolds) with cNKZ projection. Over 1000 runs with three AMMs, TSR shows significantly higher $M_{1}$ and $M_{2}$ residuals than our method (\cref{table:repcompare}), due to their reliance on the pseudo-Jacobian inverse, which introduces higher errors.

\subsection{Coordinated Constrained Motion Planning on Intersection of Manifolds}

\begin{figure}[t]
    \begin{subfigure}[t]{0.17\textwidth}
        \includegraphics[width=\textwidth, trim={0 0 0 0cm},clip]{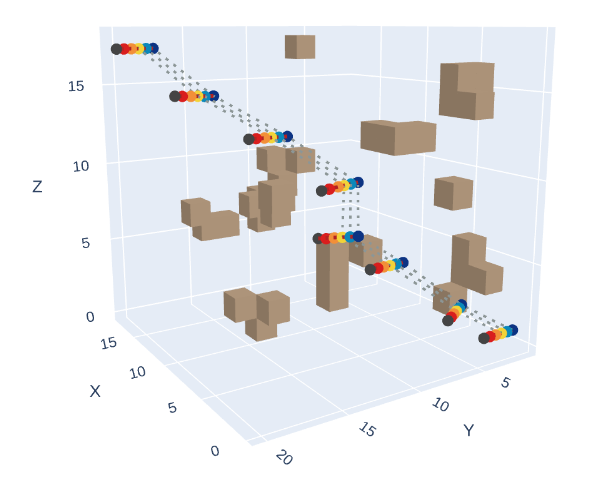}
        \caption{S\_5 (Medium)}
    \end{subfigure}
    \begin{subfigure}[t]{0.15\textwidth}
        \includegraphics[width=\textwidth, trim={0 0 0 1.5cm},clip]{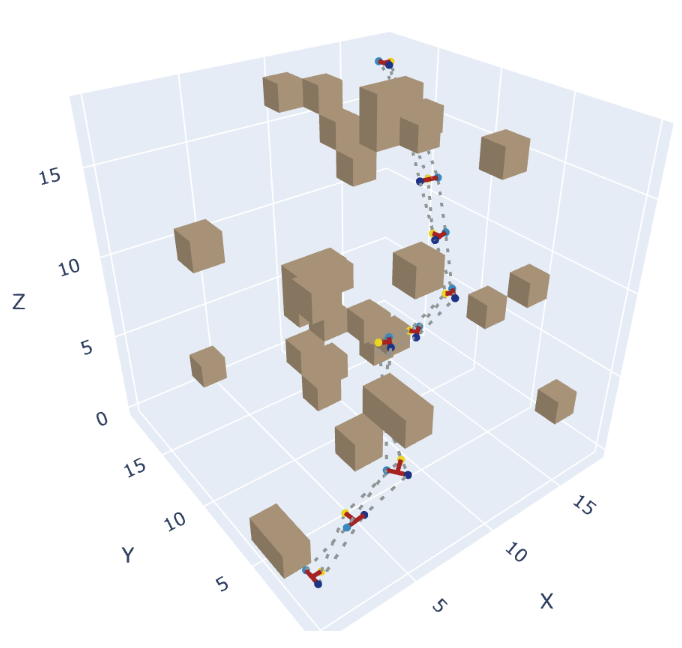}
        \caption{T\_3 (High)}
    \end{subfigure}
    \begin{subfigure}[t]{0.15\textwidth}
        \includegraphics[width=\textwidth, trim={0 0 0 0cm},clip]{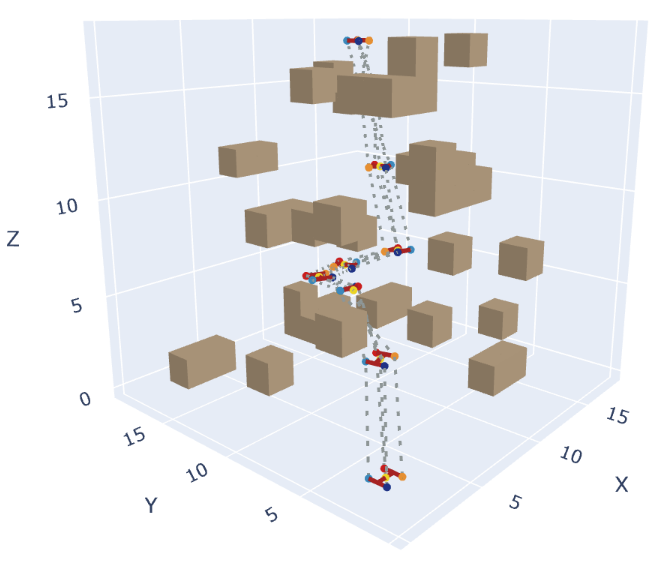}
        \caption{I\_5 (High)}
    \end{subfigure}
    \caption{Sample paths generated for 6 AMMs carrying a straight structure, 3 AMMs carrying a T-shaped structure, and 5 AMMs carrying an I-shaped structure in different complexity environments.}
  \label{figure:path}
\end{figure}

To evaluate the integration of cNKZ with RRT, we randomly generated environments of increasing complexity in simulation. We categorized the environmental complexity based on the volume ratio of free space and overall space. Thus, we generated 50 environments ($20~m~\times~20~m~\times~20~m$) of low, medium, and hard complexity each. Our task is to find a feasible path to transport a structure from an initial position to a goal position over a constrained tree. We assumed the team of AMMs was holding the structure at predefined contact points at the start of the experiments. We tested the real-world feasibility of generating executable trajectories using three non-holonomic TurtleBot3 Waffle Pi robots with OpenMANIPULATOR-X arms.

\subsubsection{Comparison: Varying Team Size and Diverse Structural Shapes} \label{sub:inc_amms}

To gauge the effectiveness of the cNKZ projection for path generation, we employ up to six AMMs (\cref{figure:path}) and evaluate on a variety of structures: straight structures with lengths ranging from $1~m$ to $2.5~m$, each with fixed contact points set at intervals of $L = 0.5~m$ from the preceding point, as well as T-shaped ($1~m~\times~1~m~\times~0.5~m$) and I-shaped ($1~m~\times~1~m~\times~1~m$) structures using teams of three and five AMMs, respectively. All contact points are illustrated in \cref{tab:planning}. Projection techniques perform optimization in the composite robot configuration space. 

Compared to RRT with NKZ projection, the cNKZ projection method demonstrated a higher success rate of up to 15\% across various complexity levels with up to $5 \times$ faster computational time for generated plans. cNKZ also has a higher successful projection rate and is up to $10\times$ faster while computing successful projections.

Incorporating the cNKZ projection method facilitated successful navigation while transporting straight structures using a six-AMM team, even in challenging complexity environments, although with a reduced success rate. Notably, the cNKZ algorithm simultaneously resolved $80$ nonlinear constraints for this experiment. 

\begin{figure}[t]
    \centering
    \includegraphics[width=8.5cm, trim={0cm 0cm 0cm 0cm}, clip]{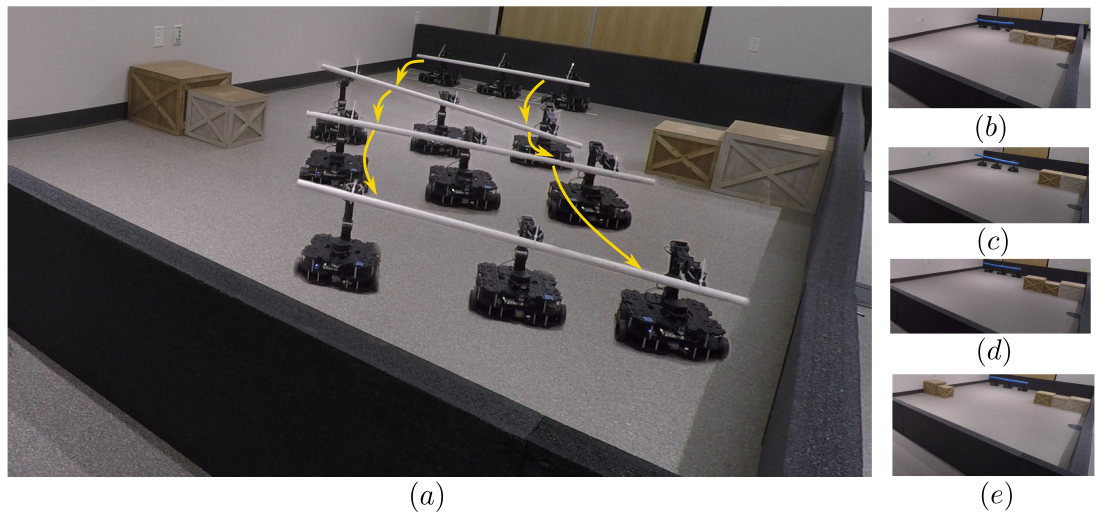}
    \caption{(a) Three robots successfully maneuver a large structure around obstacles of comparable dimensions in a congested space. (b)-(e) Four different environments that are utilized for hardware trials.}
    \label{fig:hardware}
\end{figure}

\subsubsection{Hardware Experiments}\label{sec:hardware}

Experiments used a $1.2~m$ flexible rod and a rigid PVC rod in four $3.2~m~\times~5~m$ environments with large obstacles\footnote{Video of sample trajectories from our hardware experiments is available in the supplementary materials.}~(\cref{fig:hardware}). Notably, the obstacles in these environments are comparable to the structure and the robot team, occupying more than half of the space along one dimension in the arena. This setup significantly complicates maneuvering the structure. We conducted 16 trials (8 per rod type), executing two trajectories per environment. Completing the task requires continuous satisfaction of all structural ($M_{1}$ and $M_{2}$), task-induced ($M_{3}$), and robot-specific operational ($M_{5}$) constraints. 
We plan in the combined position and velocity space of the robot, and impose a constraint ($M_{5}$) to enforce the differential drive constraint.

The robot team successfully navigated the structure in every trial, however an average execution error of $0.22 \pm 0.04 ~m$ and $12 \pm 7$ degrees is observed between the planned and actual end locations for the base robots, due to execution errors (e.g., wheel slippage).

\section{Conclusions and Future Work}
We have introduced the constrained nonlinear Kaczmarz (cNKZ) projection technique, which outperforms existing nonlinear solvers, particularly Newton-Raphson (NR) and nonlinear Cimmino's (CIM) algorithm, achieving over 90\% successful projections at twice the speed of a baseline nonlinear Kaczmarz (NKZ) approach. When integrated within a sampling-based planner, cNKZ can solve for plans for 6 AMMs (36 DoF) with up to 80 constraints, improving planning times over the baseline NKZ approach by an order of magnitude and succeeding where NR and CIM fail.
We also have introduced a framework for modeling the diverse constraints imposed on multi-robot teams as implicit manifolds, effectively representing structural, task-induced, and robot-specific operational constraints.

Although incorporating cNKZ within a sampling-based planner generated successful plans, the number of successful projections discarded due to environmental collision is substantial. In addition, given a heavy payload, it is observed that the motion of robots is heavily altered when performing the task. Recent approaches have addressed this by modeling forces and integrating it with higher level motion planner~\cite{kennel2024payload}. These caveats motivate a more efficient and scalable constrained motion planning algorithm to handle online execution, for which our proposed cNKZ method and constraint formulation lay the foundation.

\section{Acknowledgment}
{We extend our gratitude to Scott Chow (Berkshire Grey), Graeme Best (University of Technology Sydney) and Dongsik Chang (Amazon Lab126) for their insightful comments and support.}

\printbibliography{}

@article{buss2004introduction,
  title={Introduction to inverse kinematics with jacobian transpose, pseudoinverse and damped least squares methods},
  author={Buss, Samuel R},
  journal={IEEE Journal of Robotics and Automation},
  volume={17},
  number={1-19},
  pages={16},
  year={2004}
}

@ARTICLE{kennel2024payload,
  author={Kennel-Maushart, Florian and Coros, Stelian},
  journal={IEEE Robotics and Automation Letters}, 
  title={Payload-aware Trajectory Optimisation for Non-holonomic Mobile Multi-robot Manipulation with Tip-Over Avoidance}, 
  year={2024},
  pages={1-8}
}

@article{kingston2018sampling,
  title={Sampling-based methods for motion planning with constraints},
  author={Kingston, Zachary and Moll, Mark and Kavraki, Lydia E},
  journal={Annual review of control, robotics, and autonomous systems},
  volume={1},
  pages={159--185},
  year={2018},
  publisher={Annual Reviews}
}

@article{hauser2010multi,
author = {Kris Hauser and Jean-Claude Latombe},
title ={Multi-modal Motion Planning in Non-expansive Spaces},
journal = {The International Journal of Robotics Research},
volume = {29},
number = {7},
pages = {897-915},
year = {2010},
doi = {10.1177/0278364909352098},
URL = {https://doi.org/10.1177/0278364909352098},
eprint = {https://doi.org/10.1177/0278364909352098}
}

@article{berenson2011task,
  title={Task space regions: A framework for pose-constrained manipulation planning},
  author={Berenson, Dmitry and Srinivasa, Siddhartha and Kuffner, James},
  journal={The International Journal of Robotics Research},
  volume={30},
  number={12},
  pages={1435--1460},
  year={2011},
  publisher={SAGE Publications Sage UK: London, England}
}

@article{knoll2004jacobian,
  title={Jacobian-free Newton--Krylov methods: a survey of approaches and applications},
  author={Knoll, Dana A and Keyes, David E},
  journal={Journal of Computational Physics},
  volume={193},
  number={2},
  pages={357--397},
  year={2004},
  publisher={Elsevier}
}

@book{berenson2011constrained,
  title={Constrained manipulation planning},
  author={Berenson, Dmitry},
  year={2011},
  publisher={Carnegie Mellon University}
}

@article{gammell2021asymptotically,
  title={Asymptotically optimal sampling-based motion planning methods},
  author={Gammell, Jonathan D and Strub, Marlin P},
  journal={Annual Review of Control, Robotics, and Autonomous Systems},
  volume={4},
  pages={295--318},
  year={2021},
  publisher={Annual Reviews}
}

@inproceedings{gharbi2008sampling,
  title={A sampling-based path planner for dual-arm manipulation},
  author={Gharbi, Mokthar and Cort{\'e}s, Juan and Simeon, Thierry},
  booktitle={2008 IEEE/ASME International Conference on Advanced Intelligent Mechatronics},
  pages={383--388},
  year={2008},
  organization={IEEE}
}

@incollection{cortes2005sampling,
  title={Sampling-based motion planning under kinematic loop-closure constraints},
  author={Cort{\'e}s, Juan and Sim{\'e}on, Thierry},
  booktitle={Algorithmic Foundations of Robotics VI},
  pages={75--90},
  year={2005},
  publisher={Springer}
}

@article{geraerts2004comparative,
  title={A comparative study of probabilistic roadmap planners},
  author={Geraerts, Roland and Overmars, Mark H},
  journal={Algorithmic foundations of robotics V},
  pages={43--57},
  year={2004},
  publisher={Springer}
}

@article{lavalle1998rapidly,
author="LaValle, Steven",
title="Rapidly-exploring random trees : a new tool for path planning",
journal="Research Report 9811",
publisher="Department of Computer Science, Iowa State University",
year="1998",
URL="https://cir.nii.ac.jp/crid/1573950399665672960"
}

@article{hao2021gradient,
  title={A gradient descent method for solving a system of nonlinear equations},
  author={Hao, Wenrui},
  journal={Applied Mathematics Letters},
  volume={112},
  pages={106739},
  year={2021},
  publisher={Elsevier}
}

@inproceedings{ge2016rrt,
  title={RRT-GD: An efficient rapidly-exploring random tree approach with goal directionality for redundant manipulator path planning},
  author={Ge, Junxiang and Sun, Fuchun and Liu, Chunfang},
  booktitle={2016 IEEE International Conference on Robotics and Biomimetics (ROBIO)},
  pages={1983--1988},
  year={2016}
}

@article{griewank2012broyden,
  title={Broyden updating, the good and the bad},
  author={Griewank, Andreas},
  journal={Optimization Stories, Documenta Mathematica. Extra Volume: Optimization Stories},
  pages={301--315},
  year={2012}
}

@article{akram2015newton,
  title={Newton raphson method},
  author={Akram, Saba and Ann, Quarrat Ul},
  journal={International Journal of Scientific \& Engineering Research},
  volume={6},
  number={7},
  pages={1748--1752},
  year={2015}
}

@INPROCEEDINGS{kaiser2012constellation,  
author={Kaiser, Peter and Berenson, Dmitry and Vahrenkamp, Nikolaus and Asfour, Tamim and Dillmann, Rüdiger and Srinivasa, Siddhartha},  booktitle={Proc. IEEE International Conference on Robotics and Automation},
title={Constellation - An algorithm for finding robot configurations that satisfy multiple constraints},   
year={2012},  
volume={},  
number={},  
pages={436-443}}

@article{wang2022nonlinear,
title = {Nonlinear Kaczmarz algorithms and their convergence},
journal = {Journal of Computational and Applied Mathematics},
volume = {399},
pages = {113720},
year = {2022},
author = {Qifeng Wang and Weiguo Li and Wendi Bao and Xingqi Gao},
}

@INPROCEEDINGS{englert2021sampling,
	bibkey = {bib/englert2021rss.bib},
	Video = {https://www.youtube.com/watch?v=Q8kbILTRxfU},
	Code = {https://github.com/etpr/sequential-manifold-planning},
	Author = {Peter Englert and Isabel M. Rayas Fern{\'a}ndez and Ragesh Kumar Ramachandran and Gaurav S. Sukhatme},
	Booktitle = {Proc. Robotics: Science and Systems Conference},
	Title = {Sampling-Based Motion Planning on Sequenced Manifolds},
	Year = {2021}
}

@INPROCEEDINGS{berensom2009manipulation,  
author={Berenson, Dmitry and Srinivasa, Siddhartha S. and Ferguson, Dave and Kuffner, James J.},  
booktitle={Proc. IEEE International Conference on Robotics and Automation}, title={Manipulation planning on constraint manifolds},  
year={2009},  
volume={},  
number={},  
pages={625-632}}

@article{nonlinearcimmino,
author = {Iusem, A N and De Pierro, A R},
title = {Convergence Results for an Accelerated Nonlinear Cimmino Algorithm},
year = {1986},
issue_date = {Aug. 1986},
publisher = {Springer-Verlag},
address = {Berlin, Heidelberg},
volume = {49},
number = {4},
journal = {Numerical Mathematics},
month = {Aug},
pages = {367–378},
numpages = {12}
}

@inproceedings{zhang2021task,
  title={Task-space decomposed motion planning framework for multi-robot loco-manipulation},
  author={Zhang, Xiaoyu and Yan, Lei and Lam, Tin Lun and Vijayakumar, Sethu},
  booktitle={2021 IEEE International Conference on Robotics and Automation},
  year={2021}
}

\end{document}